\pdfoutput=1

\documentclass[11pt]{article}

\usepackage[]{ACL2023}
 \usepackage{import}
\usepackage{times}
\usepackage{latexsym}
\usepackage{hyperref}
\usepackage{array} 
\usepackage{rotating}
\usepackage[T1]{fontenc}

\usepackage[utf8]{inputenc}

\usepackage{microtype}

\usepackage{inconsolata}
\usepackage{graphicx}
\usepackage{booktabs}
\usepackage{multirow}
\usepackage{amsmath}
\usepackage{pifont}
\usepackage{pdflscape}
\usepackage[raggedrightboxes]{ragged2e}
\definecolor{conclusiongreen}{RGB}{1, 113, 0} 
\definecolor{reasonblue}{RGB}{0, 118, 186}

\title{STRONG -- Structure Controllable Legal Opinion Summary Generation}

\author{Yang Zhong \\
  University of Pittsburgh \\
  Pittsburgh, PA \\
  \texttt{yaz118@pitt.edu} \And
  Diane Litman \\
  University of Pittsburgh \\
   Pittsburgh, PA \\
  \texttt{dlitman@pitt.edu} \\}

\begin{document}
\maketitle
\begin{abstract}
We  propose an approach for the structure controllable summarization of long legal opinions that considers the argument structure of the document. Our approach involves 
using predicted argument role information to guide the model in generating coherent summaries that follow a provided structure pattern.
We demonstrate the effectiveness of our approach on a dataset of
legal opinions and show that it outperforms several 
strong baselines with respect to ROUGE, BERTScore, and structure similarity.

\end{abstract}
\section{Introduction}\label{sec:intro}
\begin{figure}[t!]
\small
\begin{center}
\includegraphics[width=0.41\textwidth]{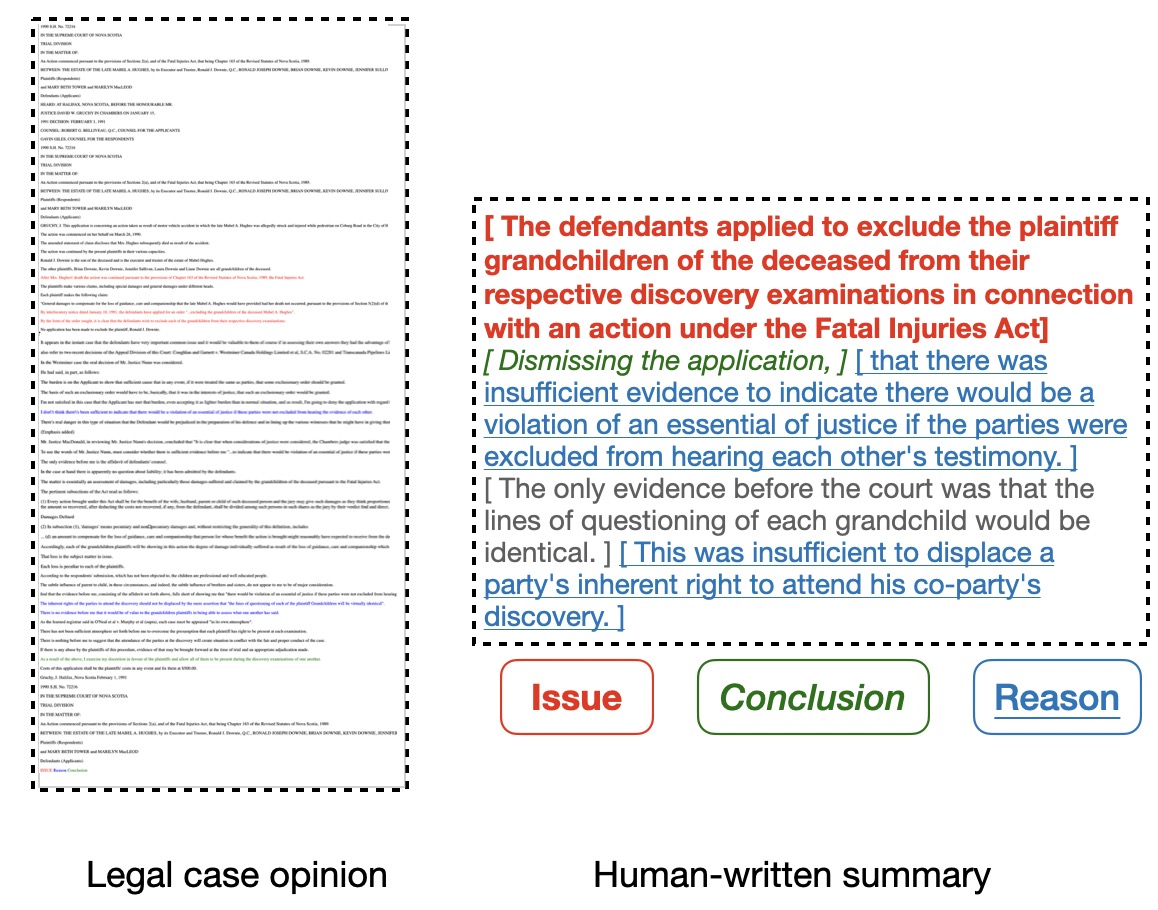}
 \caption{Example of a legal case opinion with its summary. The summary is annotated with oracle 
 argument structure labels (one \textbf{\textcolor{red}{Issue}}, one \textit{\textcolor{conclusiongreen}{Conclusion}}, and two \underline{\textcolor{reasonblue}{Reasons}}).  Presenting an issue followed by a conclusion and reasons is the dataset's most common normalized structure pattern (54\%). Complete descriptions of patterns are in Appendix \ref{appendix:structure_pattern}.}
 \label{fig:intro}
 \end{center}
\end{figure}

Discourse structure plays an essential role in text generation in domains ranging from news  
\cite{van_2013} 
to peer-reviewed articles \cite{shen-etal-2022-mred}. 
In the legal domain, it's equally important to draft a summary that can follow a blueprint \cite{xu-2021-position-case}. 
For instance, in Figure \ref{fig:intro}, given a long legal opinion with thousands of words as input, a legal expert organized the summary by making the argument clear in terms of the issues the decision addressed,  the decision's conclusion, and the reasoning behind the decision.


While progress has been made in controllable generation, limited research has 
controlled discourse structure.
Recently, \citet{Spangher2023SequentiallyCT} and \citet{Shen2022SentBSSB} proposed approaches to generate sentences with discourse structure labels. However, no existing controllable generation work addresses the legal domain, where the argumentative structure is pivotal. While prior work in the legal field highlighted the significance of argumentative structure from the input \cite{elaraby-litman-2022-arglegalsumm}, the potential for utilizing argument structure to guide text generation remains unexplored.

Based on a corpus analysis showing that experts
use common patterns to summarize legal opinions (the most frequent one is shown in Figure \ref{fig:intro}), we develop a novel structure-prompting approach called STRONG (\textbf{S}tructure con\textbf{TR}ollable legal \textbf{O}pi\textbf{N}ion summary \textbf{G}eneration). STRONG is implemented using
Longformer Encoder Decoder \cite{beltagy2020longformer} coupled with automatically created structure prompts.
Results demonstrate that STRONG outperforms summarization models without structure control and improves inference time over models with structure control from other domains. We make our models available at \url{https://github.com/cs329yangzhong/STRONG}.

\section{Related Work}
\label{sec:related}
Prior work on controllable generation \cite{hu-2017-control, goyal-durrett-2020-neural, dou-etal-2021-gsum, He2022CTRLsumTG} has focused on inner-sentence token-level attributes (e.g., syntactic structure) or full-text stylistic features (e.g., sentiment/topic). Recent research started looking at generating long texts adhering to discourse structures derived from news or article reviews \cite{ghazvininejad-etal-2022-discourse, ji-huang-2021-discodvt, Spangher2023SequentiallyCT, shen-etal-2022-mred}. 
\citet{Shen2022SentBSSB} framed the task as a sentence-by-sentence generation, 
which led to a longer inference time compared to token generation baselines. \textit{
We explore structure control in legal opinions, which is challenging due to long input texts and argumentative discourse structures.}

In the legal domain, besides directly adopting the raw document-summary pairs into supervised training using abstractive summarization models such as BART \cite{lewis2020bart} and Longformer Encoder Decoder (LED) \cite{beltagy2020longformer}, \citet{elaraby-litman-2022-arglegalsumm} proposed highlighting the salient argumentative sentences in the inputs and training a model that is argument-aware. \textit{
We instead focus on improving argument structure adherence by exploiting the summaries' annotated discourse structures to create structure prompts rather than by manipulating  the original articles.}

\section{Dataset}\label{sec:dataset}
\begin{table}[t!]
\small
    \centering
    \setlength\tabcolsep{2pt}
    \renewcommand{\arraystretch}{1}
    \begin{tabular}{l|c|c|c|c}
    
   \toprule
 \textbf{Split} & \textbf{Case/Summ pairs} &  \textbf{Case len} & \textbf{Summ len} & \textbf{sents} \\
  \midrule
   \multicolumn{5}{c}{\textit{No Manual Annotations}} \\
   \midrule 
   Train &  21794 & 3979.4 & 276.2 & 10.9 \\
   Valid & 2724 & 4067.4 & 279.8 & 11.0 \\
    
    Test & 2723 & 3899.9 & 278.8 & 10.9 \\
    \midrule 
    \multicolumn{5}{c}{\textit{Manual IRC Annotations}} \\
    \midrule 
    1049-test & 1049 & 3741.1 &  245.4 & 11.0\\
    
    \bottomrule
    \end{tabular}
    \caption{Dataset statistics of CanLII. Case/Summary len is the text length in terms of  the number of words, while sents is the sentence count per summary.}
    \label{tab:data_stats}
\end{table}

\begin{figure*}[ht!]
\small
\begin{center}

 \includegraphics[width=12cm, height=6cm]{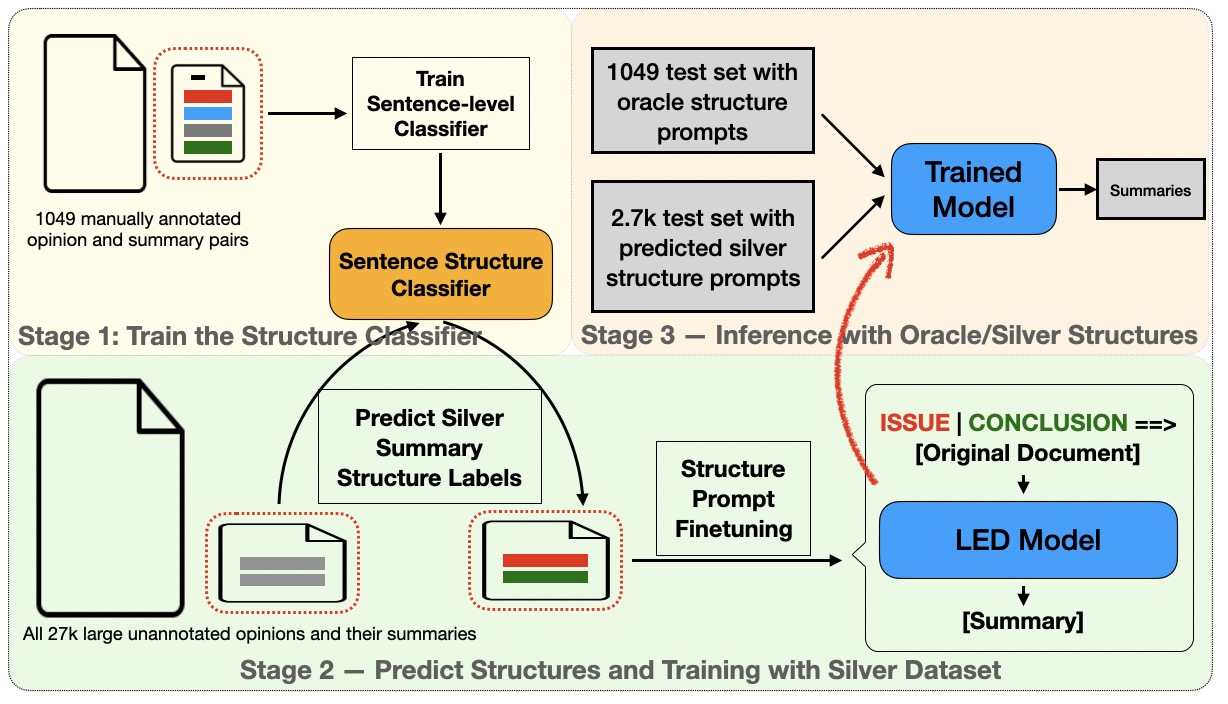}
 \caption{ Illustration of our structure prompting approach (STRONG).}
 \label{fig:structure_prompts}
 \end{center}
\end{figure*}
We leverage the CanLII dataset of legal case opinions and human-written abstractive summaries.\footnote{The data was obtained through an agreement with the
Canadian Legal Information Institute (CanLII): \url{https://www.canlii.org/en/}}\footnote{The corpus is moderately abstractive: The overlap ratios for the 1/2/3-gram between the source document and the human-authored summaries stand at 89.7\%, 62.0\%, and 42.1\%, respectively, which suggests a moderate level of abstractiveness of the dataset compared to others such as TL;DR \cite{volske-etal-2017-tl}. It can thus serves as a useful testbed for abstractive summarization.}
It consists of 28,290 legal opinions and human-written summary pairs. For testing, we first leverage the annotated subset produced by \citet{xu-2021-position-case}, including 1,049 pairs with manually annotated \textbf{IRC argument labels}: \textit{Issues} (the legal questions addressed in the case), \textit{Conclusions} (the court’s decisions for the related issue),
\textit{Reasons} (text snippets illustrating the reasons for the court's decision) and \textit{Non\_IRC} (none of the above). 
We further split the remaining 27,241  unannotated pairs into 80/10/10 percent for model training, validation, and extra testing. Corpus statistics are in Table \ref{tab:data_stats}. 

As introduced in \S \ref{sec:intro} and Figure \ref{fig:intro}, legal experts devised different strategies to construct the summaries. We thus analyze the patterns of the IRC labels in the 1,049 annotated summaries. To comprehend the high-level structures better, we remove the Non\_IRC tags and collapse adjacent text segments with the same tag into one. The most common "normalized" patterns are  ``Issue -- Conclusion -- Reason'' (54\%)  and ``Issue -- Conclusion -- Reason -- Conclusion'' 
(9\%). 
Pie charts of the top normalized and original patterns are in Appendix \ref{appendix:structure_pattern}. 

\section{Method}
Figure~\ref{fig:structure_prompts}
illustrates our proposed STRONG approach. We start by extending the small-scale annotations to the larger dataset. 
Since we only have the 1,049 test set manually annotated with oracle summary argument labels, different from \citet{elaraby-litman-2022-arglegalsumm} who used a classifier on input sentences, we propose to train a sentence classifier on summary sentences (Stage 1) and then utilize it to predict silver labels for all unannotated summaries in Stage 2.\footnote{We include the model details in Appendix \ref{appendix:irc_classifier}.} Our approach distinguishes itself from \citet{shen-etal-2022-mred}, which relied solely on manually annotated structure sequences, resulting in a smaller training set than our larger dataset with silver labels. In the next step of Stage 2, we introduce special marker tokens to guide the model in generating summaries following specified structure patterns. 
Specifically, we extract the argumentative ``IRC'' labels from  summary sentences, concatenate them with split " | " tokens and prepend before the original input text, and connect them with a special marker ``==>''. This operationalizes the argument mining of salient information blueprint, providing better guidance for the model in generating legal summaries. That is,  Stage 2 utilizes the predicted structure labels to fine-tune the LED model. 
Once the model has been trained, we generate summaries using different sets of structure labels for the two test sets during Stage 3 of the inference process.
\section{Experimental Setup}\label{sec:experimental_setup}

We compare  STRONG to two  baselines. 
\textbf{NoStructure} uses the Longformer-Encoder-Decoder (LED) base model for generating summaries. 
The second baseline re-implements \textbf{SentBS} \cite{Shen2022SentBSSB} and is structure-aware.
It uses a prompt-based backbone model to generate sentences, optimizing candidate selections based on the model likelihood and structure label probability.  All implementation details 
are in Appendix \ref{appendix:hyperparams}.

All experiments are evaluated using ROUGE-1 (R-1), ROUGE-2 (R-2), ROUGE-L
(R-L) F1 \cite{lin-2004-ROUGE},  BERTScore (BS) \cite{zhang2019bertscore}, 
and structure similarity (\textit{SS})  \cite{shen-etal-2022-mred}. 
More details on the structure metric 
are in Appendix \ref{appendix:structure_similarity}. 

\section{Results and Analysis}
\subsection{Automatic Result}

\begin{table*}[t]
\small
    \centering
    \renewcommand{\arraystretch}{0.8}
    \setlength\tabcolsep{6pt} 
    \begin{tabular}{c|l|cccc|c|cl}
    \toprule
      \textbf{ID} & \textbf{Model}  &  \textbf{R-1} & \textbf{R-2} & \textbf{R-L}   & \textbf{BS} & \textbf{SS}  & \textbf{Avg Length} & \textbf{Infer. Time}\\
         \midrule
         \multicolumn{9}{c}{{1049 Oracles}} \\
         \midrule
        \midrule
     \multicolumn{9}{l}{\small{\textit{Max output of 256 tokens}}} \\

          \midrule
        1 & SentBS & 48.31 & 23.86  & 44.73 & 86.87  & \underline{0.436}  &  129.6 & 8.5 hours\textsuperscript{\ding{169}} \\
         2 & NoStructure*  & 50.33  & 25.84  & 46.47  & 87.39 & 0.344  & 159.2 & 2.2 hours \\
           3 & STRONG* & \underline{52.47} & \underline{26.54}  & \underline{48.57}  & \underline{87.63} & 0.372 & 186.3 & 2.5 hours \\
             \midrule
             \multicolumn{9}{l}{\small{\textit{Max output of 512 tokens}}} \\
             \midrule

           4   & NoStructure & 51.61  & 26.72  & 47.76 &   87.49  &  0.383  & 198.1 & 4.2 hours \\
          
           5 & STRONG* & \textbf{55.90}   & \textbf{28.61}  & \textbf{51.97}  &  \textbf{87.78}  &  \textbf{0.535}  & 263.0 & 4.3 hours \\
           

        \midrule
        \midrule
        \multicolumn{9}{c}{{2723 Silver Test Set}} \\
        \midrule

     \multicolumn{9}{l}{\small{\textit{Max output of 256 tokens}}} \\

          \midrule
         6 & SentBS & {49.24}  & {25.43} & {45.58}  &  85.47 & 0.470   & 118.0 & 21.5 hours\textsuperscript{\ding{169}}\\
         7 & NoStructure* & 50.76  & 26.84 & 46.78  &  87.75  & 0.330  & 160.6 &6.2 hours   \\
         8 & STRONG*  &\underline{52.84}  & \underline{27.90}  & \underline{48.73} & \underline{87.97}  & \underline{0.493} & 179.3  &  6.3 hours \\
          \midrule
             \multicolumn{9}{l}{\small{\textit{Max output of 512 tokens}}} \\
             \midrule

         9 & NoStructure & 52.22  & 27.57  & 48.18  &  87.69  & 0.440   & 196.9 &  13.0 hours \\
         10 & STRONG*  & \textbf{57.17} & \textbf{29.87}  & \textbf{52.93}  &  \textbf{88.10} & \textbf{0.543}  & 255.9 & 13.1 hours \\

    \bottomrule
    \end{tabular}
    \caption{Results of different models on the CanLII oracle and silver test sets. BS refers to BERTScore, SS means structure similarity, respectively. 
    Models with * mean all results are statistically different from the previous row, based on 95\% confidence intervals. All results are reported as an average of 3 runs initialized with random seeds. Best results are highlighted with \textbf{bold}, and  best results under the 256 token settings are \underline{underlined}. Rows 1 and 6 (with \textsuperscript{\ding{169}}) experiment with an RTX3090Ti card with larger memory, which will make the inference time faster than on the default cards, which are RTX5000s and used for all other experiments.}
    \label{tab:mian_results}
\end{table*}

This section addresses two research questions: \textbf{RQ1}. Does STRONG improve summarization quality compared to baselines?  \textbf{RQ2}.  How do models compare in preserving structure? We then conduct analyses based on the observations and perform a small-scale human evaluation.\\
\textbf{RQ1.}
Using the left results section of Table \ref{tab:mian_results}, we first compare STRONG with the NoStructure  baseline on traditional ROUGE and BERTScore summarization metrics. 
For the 1049 test set, when the maximum generation output length is limited to 256 tokens, 
we observe that STRONG 
obtains an average of 2.1, 0.7, 2.1, and 0.2 improvements across ROUGE-1, 2, L, and BERTScore (rows 3 vs. 2), which are \textbf{significant} based on 95\% confidence intervals. STRONG also outperformed the re-implemented SentBS baseline (rows 3 vs. 1). We also explored the impact of increasing the maximum output length to 512 tokens,  
based on the observation that oracle summaries tended to be longer (Table \ref{tab:data_stats}). 
Similar trends were seen when the maximum output length is increased to 512 tokens (rows 5 vs. 4), as well as when all analyses are repeated using the 2,723 silver set (rows 6-8, 9-10). This illustrates that the target structure information helps STRONG generate higher-quality summaries. 
Appendices \ref{appendix:ROUGE_study_length} and \ref{appendix:examples} present examples and analysis to demonstrate model output differences in content coverage.\\
\textbf{RQ2.}\label{sec:structure_similarity}
 In the 1049 test set, compared to the NoStructure model (row 2), 
 the STRONG model (row 3) significantly improves the structure similarity scores by 0.03.
 While SentBS (row 1) outperforms both methods (rows 2/3), the tradeoff is increasing inference time (last column). 
 In contrast, with the extended 512 generation length where we could not even run SentBS, STRONG obtained the best oracle test set performance in the table, with a margin of 0.1 compared to SentBS (rows 5 vs. 1). Albeit imperfect, 
 on the silver test set where our IRC sentence classifier predicts the structure labels, STRONG also gains 0.1 improvements to NoStructure (rows 7 vs. 8, and 9 vs. 10), and now even surpasses  SentBS (row 6 vs. 8) on structure similarity while again 
reducing inference time.\\

\subsection{Length Control}
\begin{table}[htbp]
    \centering
    \small
    \setlength{\tabcolsep}{2pt}
    \begin{tabular}{lccccc}
        \toprule
        \textbf{Model} & \textbf{Control Len.} & \textbf{R-1} & \textbf{R-2} & \textbf{R-L} & \textbf{BS} \\
        \midrule
        
        NoStructure & No & 50.33 & 25.84 & 46.47 &87.39 \\
        STRONG &  No & 52.47 & 26.54 & 48.57 & 87.63\\
        \midrule
        NoStructure & Yes & 50.74 &	25.91 & 47.07 &  87.17\\
        STRONG &  Yes & 50.96 &	26.26 & 47.33 & 87.39\\
        \bottomrule
    \end{tabular}
    \caption{Results of models when summary has a maximum (top) versus controlled (bottom) length of 256 tokens.  Although STRONG still outperforms the baseline, the delta is reduced when the length is controlled.}
    \label{tab:effects_length_control}
\end{table}
The second to last column of Table \ref{tab:mian_results} shows that STRONG generates the longest summaries, which may have impacted the above assessments. 
We thus force NoStructure and STRONG to continue generating tokens until reaching the same specified limit of \{64, 128, 256, and 512\} tokens.\footnote{The generation length of SentBS cannot be rigidly regulated, considering that it adheres to a sentence-by-sentence generation paradigm, and the inconsistencies in the length of structural prompts result in diverse outputs.}
Table \ref{tab:effects_length_control} shows the results for the 256 token limit,\footnote{An analysis of additional lengths is in Appendix \ref{appendix:control_length_analysis}.}
and indicates that the Table~\ref{tab:mian_results} performance gap  (repeated in the first two rows of Table~\ref{tab:effects_length_control}) diminishes when the length is controlled (the last two rows).
This suggests that the structural benefits of STRONG become less important when output length is fixed. However, controlled length can lead to incomplete generations (see an example in Appendix \ref{appendix:control_length_analysis}), and STRONG can dynamically adjust and generate similar length summaries compared to the oracle when they can stop generation if needed. Additionally, for both NoStructure and STRONG, we observe a drop in ROUGE performance for extremely long summaries (512 tokens) compared to smaller output lengths (see Appendix \ref{appendix:control_length_analysis}), likely because 512 tokens deviate from the distribution of human summarization lengths. We additionally experimented with another setup to adjust the minimum generation length of each model and with higher length penalties.  These results are detailed in Table \ref{tab:min_longer_summary_table}, located in Appendix \ref{appendix:control_min_length_analysis}. We observed that our STRONG model outperformed the baseline and reinforced the notion that structural information plays a crucial role in guiding the model to produce summaries with the appropriate length and level of detail.\\ 
\begin{table}[htbp]
    \centering
    \small
    \begin{tabular}{l>{\hspace{1em}}l}
        \toprule
        \textbf{Model} & \textbf{\textsc{SummaC\textsubscript{Conv}}}  \\
        \midrule
        \multicolumn{2}{l}{\textit{Max output of 256 tokens}}\\
        \midrule
       SentBS &  0.660 \\
        NoStructure & 0.663\\
        STRONG &   0.704* \\
         \midrule
        \multicolumn{2}{l}{\textit{Max output of 512 tokens}}\\
        \midrule
      
        NoStructure & 0.658\\
        STRONG &    0.697*\\

        \bottomrule
    \end{tabular}
    \caption{Results of the average factuality scores for models in Table \ref{tab:mian_results} over the CanLII oracle test set. * means the result is significantly different from the previous row using paired t-test.}
    \label{tab:factuality_score}
\end{table}
\subsection{Factuality} 
 To evaluate the factuality of generated text, we picked the \textsc{SummaC\textsubscript{Conv}} score from \citet{laban-etal-2022-summac}, which utilizes the NLI model to detect summary inconsistencies and performs well on multiple factuality benchmarks (details in the original paper) compared to other metrics such as FactCC \cite{kryscinski-etal-2020-evaluating} and DAE \cite{goyal-durrett-2020-evaluating}. As shown in Table \ref{tab:factuality_score}, our STRONG model obtains the highest scores, which means the highest consistency between document and generated summaries.\\

\subsection{Human Evaluation}
Human evaluation is 
under-explored for legal 
tasks, as it is labor-intensive due to long documents / summaries 
and requires evaluators 
with legal expertise \cite{JAIN2021100388}. 
As a first step, we conducted a small-scale human evaluation using five legal decisions to assess the quality of summaries generated by all models in Table \ref{tab:mian_results}. 
Three legal experts 
were asked to evaluate the coherence of the generated texts and assess the coverage of argumentative components when compared to the oracle summaries crafted by the human CanLII experts.\footnote{We provide the evaluation details in Appendix \ref{appendix:human_evaluate}.} The evaluator feedback indicated that longer summaries could potentially introduce more factual errors, and there was inconsistency in terms of fluency and readability, with mixed performance observed (one annotator reported issues in two cases). On the other hand, the advantage of controllable structure generation was more evident when generating longer summaries. In two out of five cases, the summaries generated by STRONG  were preferred in the 512-length setting, while under the 256-length setting, only one STRONG-generated summary was favored.
\section{Conclusion}
We proposed the STRONG approach for improving the summarization of long legal opinions by providing target-side structure information. STRONG accepts different types of prompts and generates summaries accordingly. Experiments demonstrated that the content coverage, summary length, structure adherence, and inference time are all improved with STRONG compared to prior  structure-control and no-structure baselines.
\section*{Limitations}
Our research results are constrained  by our dependence on a single dataset for experimentation as well as by computing resource limitations.  While prior work demonstrated that the SentBS approach could obtain negligible performance drop with regard to automatic metrics such as ROUGE and BERTScore compared to a finetuning structure prompted baseline, our current experiment is hindered by extreme demand of GPU memories given the much longer legal input and large parameter searching space. We also demonstrate that the slowness of compared work is more severe when transferring the model to our tasks. Further experiments on more extensive setups of the prior baselines can be important for future work to verify the past work's conclusions.  We recognize that our methodology relies on annotated data for structure labels, particularly when adapting to novel domains. In future research, we aim to investigate zero-shot learning techniques to enable structure classification without the necessity for annotations.

While our paper uses standard summarization metrics and a similarity measure particularly related to our focus on structure controllability, we do not yet extensively investigate how STRONG impacts factuality besides the \textsc{SummaC\textsubscript{Conv}} score \cite{laban-etal-2022-summac}. A recent study \cite{wan-etal-2023-faithfulness} demonstrates that improvements in factuality-related metrics come with the sacrifice of dropping automatic metrics such as ROUGE and BERTScore, while \citet{min2023factscore} harness the power of LLMs to evaluate the factuality of long-form text generation. Deviating from prior work \cite{zhong2022computing} that studies the extractive summarization task, we focused on the abstractive summarization, which has shown to surpass the performance of extractive methods by a noticeable margin, while both strategies introduce unfaithfulness \cite{zhang2023extractive}.
Another limitation is that we only exploited the IRC structure representations due to the availability of oracle summary annotations.
Exploring the use of
structures based on other methods such as \citet{lu-etal-2018-object} is a promising area for future work.  Also, the automatic evaluation metrics may be deficient compared to human evaluations, thus unfaithfully representing the final quality of generated summaries compared to real legal experts. Moreover, in a real application, end users may propose and inquire about different outputs with self-designed structure prompts\footnote{We provide an example of feeding different prompts to generate diverse summaries in Appendix \ref{appendix:diff_prompt}.}, which remains an open-ended challenge and may need human validation for future works.

\section*{Ethical Considerations}
Using  generated abstractive summary results from legal opinions remains a problem, as abstractive summarization models have been found to contain hallucinated artifacts that do not faithfully present the source texts \cite{kryscinski-etal-2019-neural, zhao-etal-2020-reducing}. The generation results of our models may carry certain levels of non-factual information and need to be used with extra care. Similarly, CanLII has taken
measures (i.e., blocking search indexing) to limit the disclosure of defendants’ identities, while abstractive approaches may cause potential user information leakage. 

\section*{Acknowledgements}
This work is supported by the National Science Foundation under Grant No. 2040490 and by Amazon. We want to thank the members of the Pitt AI Fairness and Law Project members, the Pitt PETAL group, and anonymous reviewers for their valuable comments in improving this work.

\bibliography{anthology,main}
\bibliographystyle{acl_natbib}

\appendix

\newpage
\section{IRC Structure Patterns}\label{appendix:structure_pattern}
\begin{figure}[h]
\small
\begin{center}
\includegraphics[width=0.44\textwidth]{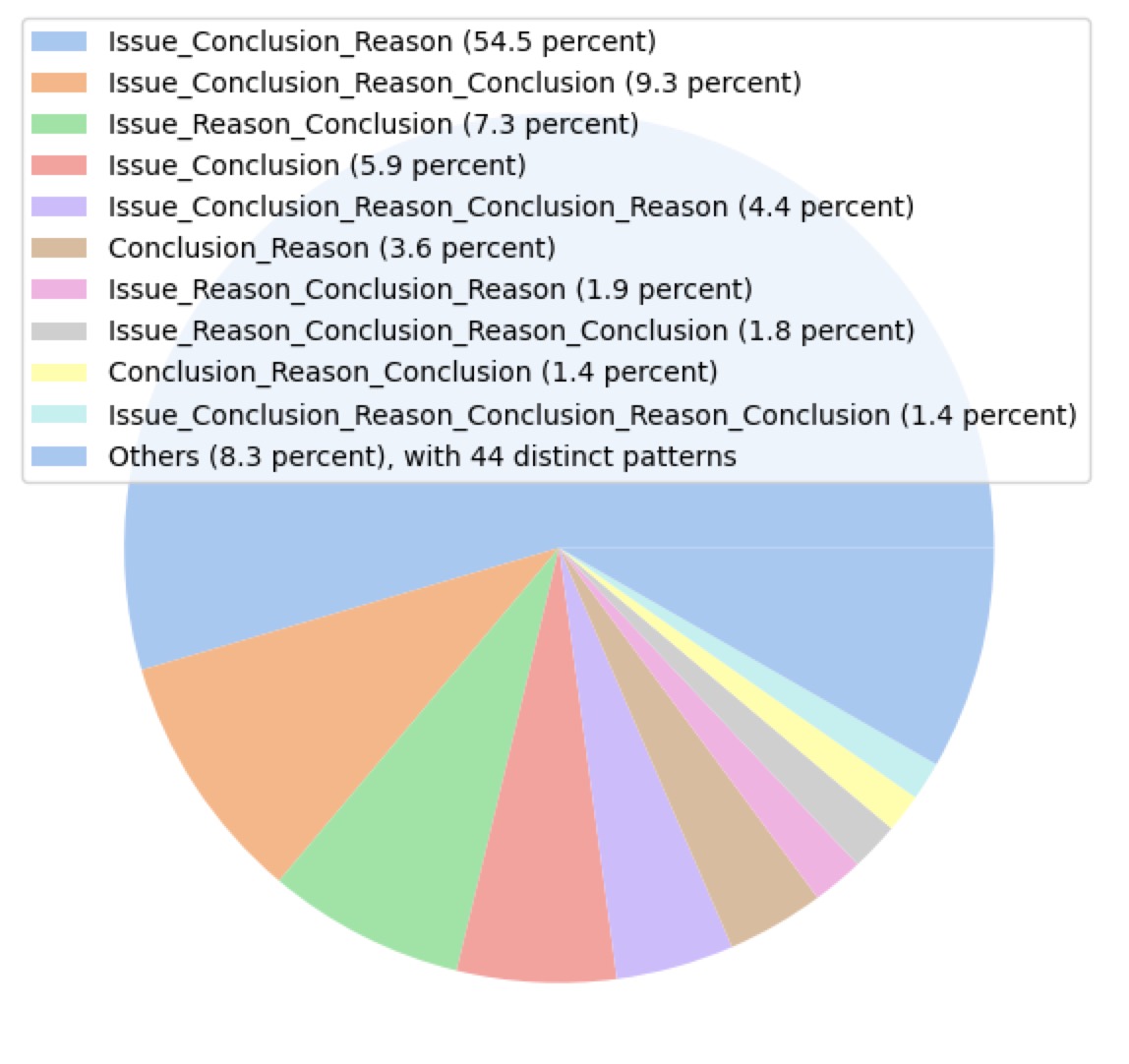}
 \caption{Pattern distribution of normalized summary structures, here we exclude the Non-IRC labels.}
 \label{fig:IRC_patterns}
 \end{center}
\end{figure}

\begin{figure}[]
\small
\begin{center}
\includegraphics[width=0.40\textwidth]{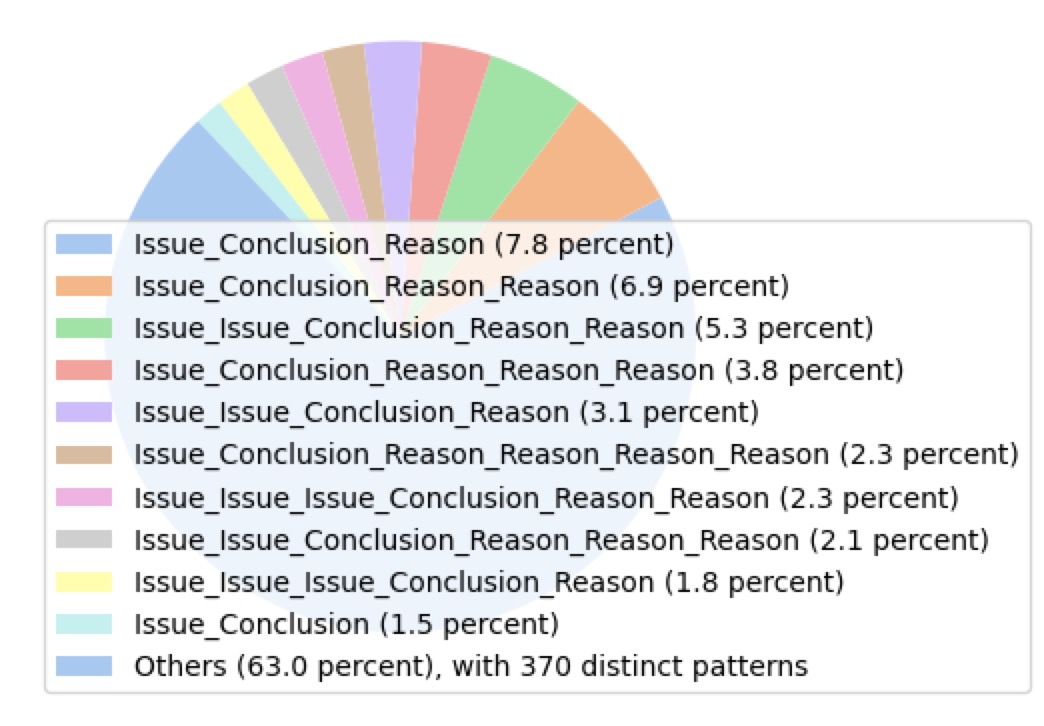}
 \caption{Pattern distribution of summary structures, here we exclude the Non-IRC labels.}
 \label{fig:IRC_patterns2}
 \end{center}
\end{figure}

We report the distribution of different structure patterns with the normalized version (we remove the neighboring duplicated labels and ignore the Non\_IRCs for better structure presentation) in Figure \ref{fig:IRC_patterns}. We observe that most 1049 test summaries are annotated in an Issue -- Conclusion -- Reasoning pattern, while the remaining have different reordering of latter patterns. Legal experts sometimes employ the ``Conclusion then Reasoning'' pattern (3.6\%) to strengthen the validity of the case summary. We found 54 distinct normalized structure patterns without considering the Non\_IRCs and varying numbers of neighboring sentences. This suggests that legal experts employed diverse strategies to construct the summaries and confirms the importance of structure modeling in text generation tasks. Regarding the original patterns (excluding Non\_IRCs), as shown in Figure \ref{fig:IRC_patterns2}, the numbers of Issue and Reasoning sentences varied.

\section{Implementation Details}\label{appendix:hyperparams}
All of our BART-based experiments and the sentence classification model are conducted on Quadro
RTX 5000 GPUs, each with 16 GB RAM. For SentBS models, we adopted the authors' original codebase workflow\footnote{\url{https://github.com/Shen-Chenhui/SentBS}} and reimplemented it on an RTX 3090Ti GPU to satisfy the minimum RAM requirements. 
\begin{figure}[t!]
\small
\begin{center}
\includegraphics[width=0.5\textwidth]{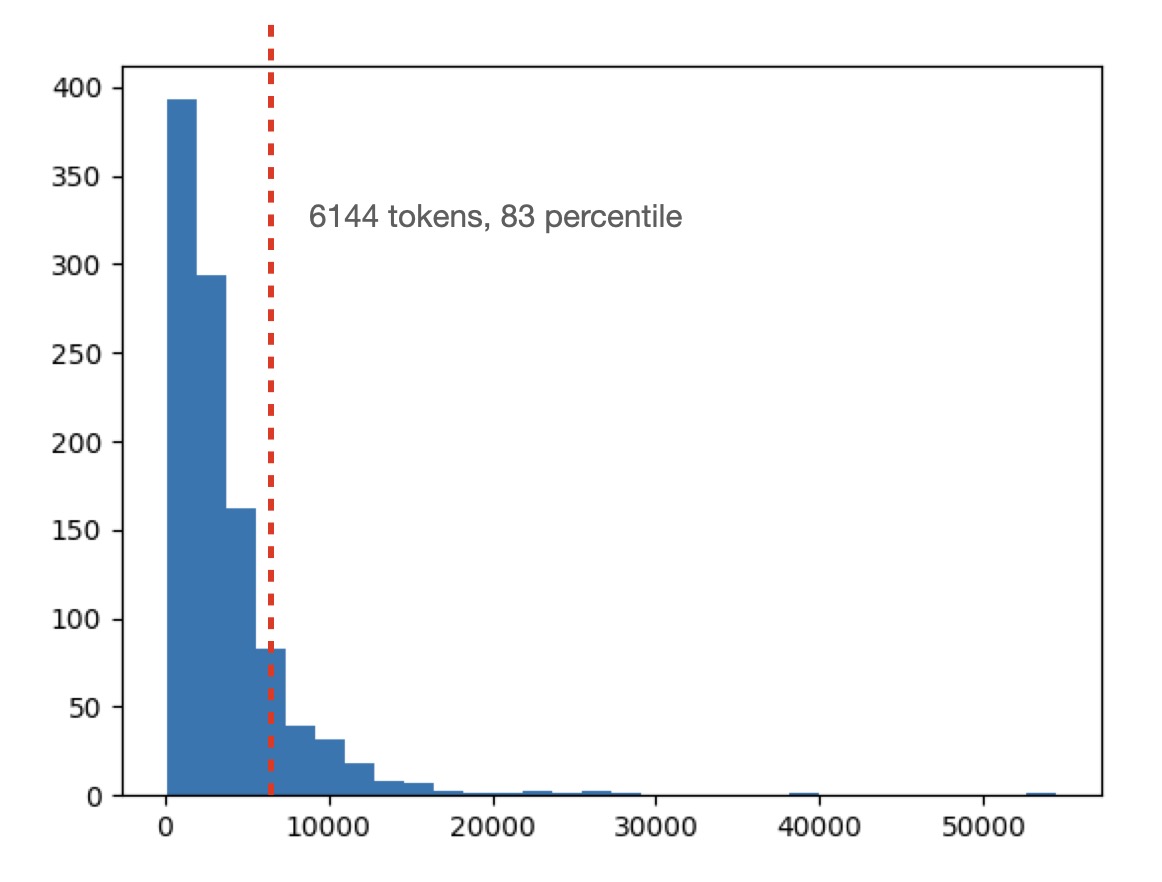}
 \caption{Input case length distribution of the 1049 test set, for models truncated at 6144 tokens, we retain 83 percent complete inputs.}
 \label{fig:length_6144}
 \end{center}
\end{figure}
\subsection{NoStructure and STRONG model}\label{appendix:baseline_setup}
All models are implemented with the Huggingface library \cite{wolf-etal-2020-transformers} using PyTorch, initialized with the \textit{"allenai/led-base-16384"} checkpoint\footnote{\url{https://huggingface.co/allenai/led-base-16384/tree/main}}.  We train all our models with the same learning rate of 2e$-$5. We train the models for 16k steps, using the gradient step of 4, batch size of 1, and save the
best checkpoints at every 1,000 steps, based on the ROUGE-2 F1 score of the validation set evaluations. Each model is trained with three randomized seeds, and we report the final averaged results. 
For training summarization models, we set the min/maximum inference summary length to 64/256 tokens. We employed beam-search with a beam size of 4 for all experiments. We additionally experimented with 512 output lengths in the main results. We truncate the input length to 6,144 tokens for
the LED-base model due to our GPU limitation, and analyze the effects of contents truncations.\footnote{We plot the length distribution of input documents in Figure \ref{fig:length_6144}.} For inference, we do a batch decoding with a batch size of 5 and report the total inference time accordingly.

\subsection{IRC Classifier Training}
\label{appendix:irc_classifier}
 Our argument role (IRC) classifier leverages a finetuned \textit{legalBERT} \cite{zhengguha2021} model due to its performance gain compared to other contextualized models such as BERT \cite{devlin-etal-2019-bert} and ROBERTa \cite{liu2019roberta} as shown in \citet{elaraby-litman-2022-arglegalsumm} to predict sentence IRC labels as a four-way classification task. We implemented the model with the PyTorch Lightning framework\footnote{\url{https://github.com/Lightning-AI/lightning}}. We split sentences from the 1049 annotated summaries into a 80/10/10 randomized setting for train, validation, and testing. For model training, we set the learning rate of 2e$-$5, training for 15 epochs, and leveraged the validation loss for early stopping criteria with a patience of 5. The final prediction  macro-F1 is 0.7586. The detailed sentence classifier result is shown in Table \ref{tab:irc_classifier}. 

\begin{table}[t!]
\small
    \centering
    \setlength\tabcolsep{3.5pt}
    \renewcommand{\arraystretch}{1}
    \begin{tabular}{l|c|c|c|c|c}
    
   \toprule

  \textbf{Data Split} & \textbf{I-F$_{1}$} & \textbf{R-F$_{1}$}  & \textbf{C-F$_{1}$}  & \textbf{Non-F$_{1}$}  & \textbf{Macro F$_{1}$}  \\
  \midrule
  
   Valid & 76.7 & 66.3 & 76.7 & 76.0 & 73.8 \\
    \midrule
    Test & 75.3 & 71.8 & 81.0 & 76.7 & 75.9 \\

    \bottomrule
    \end{tabular}
    \caption{IRC label classifier performance on the 1049 subset's validation and test split.}
    \label{tab:irc_classifier}
\end{table}
\begin{table}[t!]
\small
    \centering
    \setlength\tabcolsep{2pt}
    \renewcommand{\arraystretch}{1}
    \begin{tabular}{l|c|c|c|c|c}
    
   \toprule

  & \textbf{R-1} & \textbf{R-2}  & \textbf{R-L}  & \textbf{Infer. time} & \textbf{Avg length}   \\
  \midrule
  
    sentence-ctrl & 48.31 & 23.86 & 44.73 & 8.5 hours & 129.6\\
    segment-ctrl & 42.79 & 21.56 & 39.59 & 6 hours & 77.7 \\

    \bottomrule
    \end{tabular}
    \caption{SentBS results with different structure sequences.}
    \label{tab:sent-ctrl_vs_seg-ctrl}
\end{table}

 \subsection{SentBS Re-Implementation}
 \label{appendix:sentbs_setup}
 The original SentBS \cite{Shen2022SentBSSB} approach is implemented with a backbone of the BART-large \cite{lewis2020bart} model and using a V100 Graphic Card with 32GB memory. We first replaced the BART-large backbone with our trained LED models. Due to the limitation of GPU memory, the model failed to load on our prior RTX 5000 GPUs with the basic setting of beam size of 2. We instead ran the model on a GTX 3090Ti card with 24 GB memory, inference with the SentBS's ``beam search + nucleus sampling'' option, generation size of 4, beam size of 2, a top-p ratio at 0.9, and the maximum decoding length of 256 tokens. All other parameters are consistent with the original article experiment. Besides the sentence label searching, we additionally experiment with the segment-ctrl setup, where the target summary labels are de-duplicated to spans with non-repeated IRC labels. The results are shown in Table \ref{tab:sent-ctrl_vs_seg-ctrl}. We tested the model's performance on the original MReD dataset, which gives 34.77/9.69/30.99 regarding ROUGE scores, which is comparable to the original paper's result 34.61/9.96/30.87 with our evaluation script.

\section{Structure Similarity Evaluations}\label{appendix:structure_similarity}
As mentioned in \S \ref{sec:experimental_setup}, we adopted a metric from the human evaluation introduced in \citet{shen-etal-2022-mred} to measure the structure-similarity between a system output summary and a given oracle summary with the oracle structure prompt. In our actual implementation, the similarity score is computed by \[1 - (\dfrac{minimum\_edit\_distance(S_i, O_i)} {max(len(S_i), len(O_i))}
\] where the edit distance is computed as the Levenshtein Distance, with equal penalties for replace, insert, and delete operations. We report the average similarity score of the test sets in the table results. 
Given that the sentence classification model can make wrong predictions, we estimate an upper bound by making predictions of the human-written summary sentences, which resulted in 0.781
 for the original similarity score. Albeit not perfect, we can still assume that a generation model performs better on the structure-controlled generation task if the computed similarity becomes higher.


\newpage
\section{More Analysis on the Generation}\label{appendix:ROUGE_study_length}

\subsection{Controlled Length}\label{appendix:control_length_analysis}
\begin{figure}[th]
\small
\begin{center}
\includegraphics[width=0.46\textwidth]{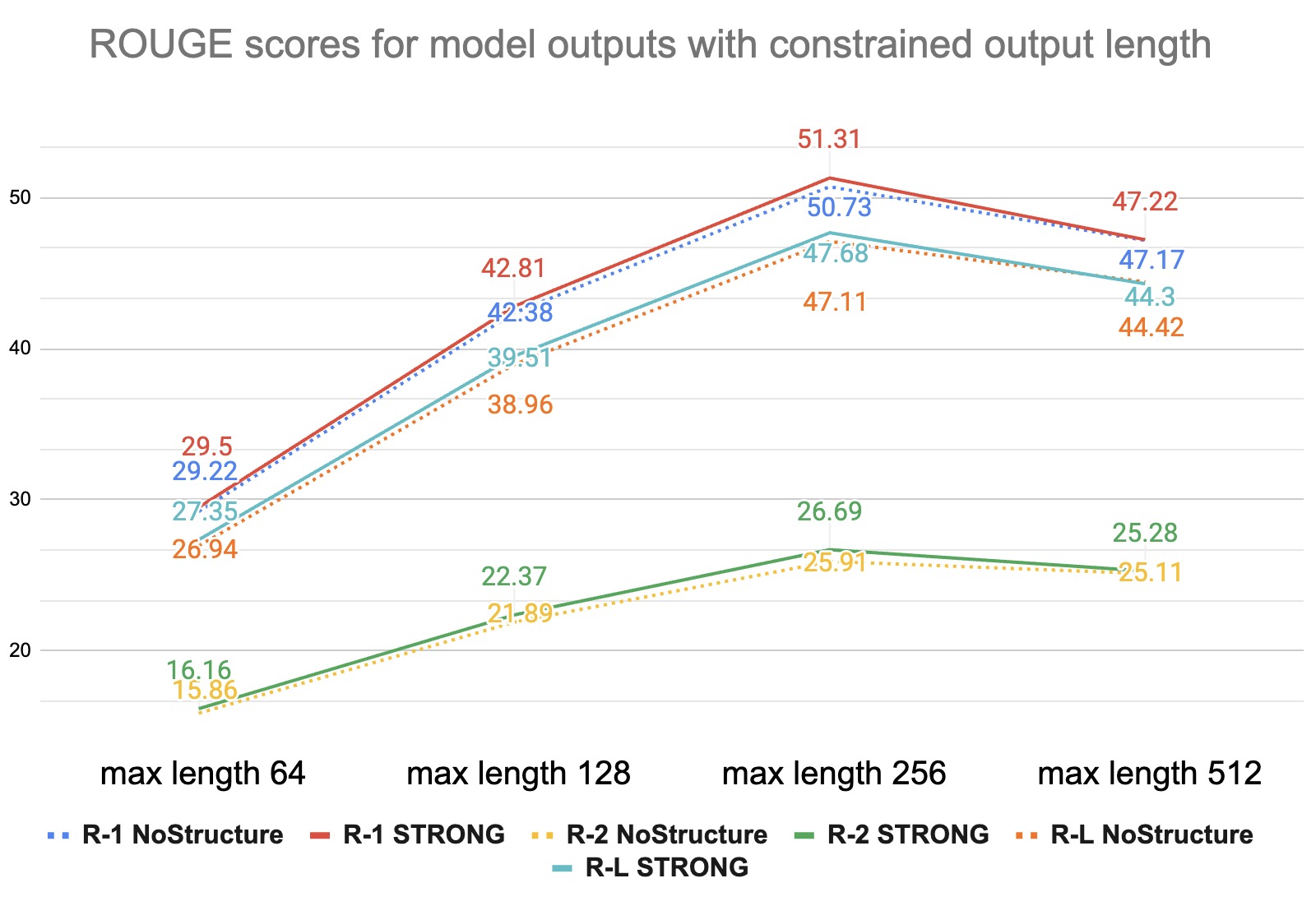}
 \caption{ROUGE scores for NoStructure and STRONG
models with 64, 128, 256, and 512 output token limits.}
 \label{fig:constrained_length}
 \end{center}
\end{figure}

We compare the ROUGE scores between the NoStructure and STRONG models and visualize the results in Figure \ref{fig:constrained_length}. The findings suggest that the performance gap between the models diminishes, indicating that the structural benefits of the STRONG model for summary organization become less significant when the output length is fixed. Additionally, we observed a drop in performance for extremely long summaries (512 tokens) as they deviated from the distribution of human summarization lengths. The BERTScore performance is shown in Table \ref{tab:bertscore_control_len}, where we observe a similar trend. 

However, controlled length can lead to the incomplete generation problem, as the model can not stop generation until it hits the desired token limit. As shown in Table \ref{tab:max_vs_control_example}, models obtain incomplete last sentences under the controlled length setting. 
\begin{table*}[hp!]
    \centering
    \small
    \begin{tabular}{ccccccc}
        \toprule
        \multicolumn{1}{c}{\textbf{Length}} & \multicolumn{3}{c}{\textbf{noStructure}} & \multicolumn{3}{c}{\textbf{STRONG}} \\
        \cline{2-4} \cline{5-7} \\
        & \textbf{BS - P} & \textbf{BS - R} & \textbf{BS - F1} & \textbf{BS - P} & \textbf{BS - R} & \textbf{BS - F1} \\
        \midrule
        64 &  89.08 & 83.36 & 86.09  &	89.22	& 83.38 & 86.17 \\
         128 & 88.27	& 85.64	& 86.91 & 88.47	& 85.67 &	87.02\\ 
         256 & 86.79 & 	87.50	& 87.17 & 87.04	& 87.80 & 	87.39\\ 
          512 & 84.56 &	88.49 &	86.46& 84.62	& 88.86	& 86.66\\
        \bottomrule
    \end{tabular}
    \caption{Evaluation of models under Controlled Length, BS - P, BS - R, and BS - F1 denote BERTScore for Precision, Recall, and F1-Score, respectively. The table presents the evaluation results of models under different controlled lengths. There still exists difference between the two models, while overall the 512 length generation becomes worse.}\label{tab:bertscore_control_len}
    \smallskip
    \raggedright
    
\end{table*}

\begin{table*}
\centering
\begin{tabular}{p{0.8in}|p{5.1in}}
\toprule
\RaggedRight{Model} & \RaggedRight{Generated Summary} \\
\midrule

\RaggedRight{Reference} & 
\RaggedRight{The appellant was convicted of indecent assault against two young girls. He appealed on five grounds related to a substantial conflict in the evidence. Dismissing the appeal, that there was no error on the part of the trial judge in weighing the evidence.} \\
\midrule
\midrule
\RaggedRight{NoStructure  \textbf{max}} &
\RaggedRight{The appellant was convicted of two counts of indecent assault against two girls, aged 13 and 16.
He was sentenced to nine months imprisonment, to be followed by two years probation.
The appellant appealed.Dismissing the appeal, that there was no error on the part of the trial judge in conducting the trial or in weighing the evidence.
After carefully reviewing the evidence, the verdict was not unreasonable or not supported by the evidence and the appeal was dismissed.} \\
\midrule
\RaggedRight{NoStructure  \textbf{controlled}} &
\RaggedRight{The appellant was convicted of two counts of indecent assault against two girls, aged 13 and 17, respectively.
He was sentenced to nine months imprisonment with respect to the first assault, followed by two years probation.
The appellant appealed.Dismissing the appeal, that there was no error on the part of the trial judge in conducting the trial or in weighing the evidence.
......
There was a substantial conflict in the evidence as to the appellant's guilt, \textbf{and he had been sentenced to 9 months imprisonment for the assault on the complainant, to}} \\
\midrule
\RaggedRight{STRONG  \textbf{max}} &
\RaggedRight{The appellant was convicted of two counts of indecent assault against two young girls.
He was sentenced to nine months imprisonment with respect to the first assault, to be followed by two years probation, and one month consecutive for the second assault.
The appellant appealed.Dismissing the appeal, that there was no error on the part of the trial judge in conducting the trial or in weighing the evidence.
After carefully reviewing the evidence, the court could not say that the verdict was unreasonable or not supported by the evidence and the appeal was dismissed.} \\
\midrule 
\RaggedRight{STRONG \textbf{controlled}} &
\RaggedRight{The appellant was convicted of two counts of indecent assault against two girls.
He was sentenced to nine months imprisonment with respect to the first count and two years probation on the second count.
The appellant appealed both convictions.Dismissing the appeal, that there was no error on the part of the trial judge in conducting the trial or in weighing the evidence.
...... \textbf{as the evidence did not support the appellant's contention that the assault was committed in bad faith and that the appellant had committed the second offence in good faith and in}} \\
\bottomrule
\end{tabular}
\caption{A sample of 256 token generation for NoStructure and STRONG models under the max and control length settings. \textbf{Bold} sentences are incomplete under the controlled length setting. }
\label{tab:max_vs_control_example}
\end{table*}

\subsection{Controlled Min Length}\label{appendix:control_min_length_analysis}

 We set the minimum length parameter of the generation to (64, 128, 256, and 512) and fixed the maximum length at 512. We modified the length penalty to 2.0, aiming to prompt the model to generate longer sequences. Table \ref{tab:min_longer_summary_table} indicates that our approach yields summaries with higher ROUGE scores when a larger length penalty is applied. This positive impact remains consistent even when we set the minimum length to less than 256 tokens. These findings reinforce the notion that structural information plays a crucial role in guiding the model to produce summaries with the appropriate length and level of detail. Interestingly, in the extreme scenario where we set the minimum tokens to 512, both models perform similarly.

 \begin{table*}[tp]
    \centering
    \begin{tabular}{cccc|ccc}
        \toprule
        \multicolumn{1}{c}{\textbf{Length}} & \multicolumn{3}{c}{\textbf{noStructure}} & \multicolumn{3}{c}{\textbf{STRONG}} \\
        \cline{2-4} \cline{5-7} \\
        & \textbf{R-1} & \textbf{R-2} & \textbf{R-L} & \textbf{R-1} & \textbf{R-2} & \textbf{R-L} \\
        \midrule
        64 &  
52.00
 & 26.82
 & 48.19

&	
55.68
& 28.30
& 51.74
\\
 128 & 
 52.42
& 26.97
& 48.61
& 55.51
& 28.22
& 51.62
\\ 
         256 & 
52.30
& 26.92
& 48.73
& 53.88
& 27.65
& 50.22

\\ 
 512 &         
47.09
&24.98
& 44.32
& 46.96
& 24.95
&44.04

\\
        \bottomrule
    \end{tabular}
    \caption{Evaluation of models under Controlled Minimum Length. The table presents the evaluation results of models under different controlled minimum lengths. A difference still exists between the two models, while overall, the 512 length generation becomes worse.}\label{tab:min_longer_summary_table}
    \smallskip
    \raggedright
    
\end{table*}

\subsection{Complete ROUGE scores}
To evaluate the advantages brought by the proposed methods, alongside diagnosing the effects of augmenting the maximum generation length, we report the complete ROUGE scores of the models on the 1049 test set in Table \ref{tab:complete_results_appendix}.  Initial observations highlight that the incorporation of structural information fosters enhancements in ROUGE recall scores, despite inducing a slight decrement in precision (as evidenced in row 2/3 and row 4/5).  Additionally, the expansion of maximum output length significantly boosts the ROUGE recall, which can be attributed to the coverage of more n-grams. However, a corresponding decline in the precision score has been observed. This observation echoed with the preliminary human evaluation, which suggested that the longer outputs occasionally encompassed with higher error rate of contents, thus having lower quality.
\begin{sidewaystable}[t]
\small

    \centering
    \renewcommand{\arraystretch}{0.8}
    \setlength\tabcolsep{6pt} 
    \begin{tabular}{c|l|ccc|ccc|ccc}
    \toprule
      ID & Model  &  R-1 Precision &  R-1 Recall & R-1 F1 &  R-2 Precision & R-2 Recall & R-2 F1 &  R-L Precision & R-L Recall & R-L F1 \\
         \midrule
         \multicolumn{10}{c}{{1049 Oracles}} \\
         \midrule
        \midrule
     \multicolumn{10}{l}{\small{\textit{Max output of 256 token}}} \\

          \midrule
        1 & SentBS & 59.93 & 44.93 & 48.27 & 29.75 & 22.10 & 23.80 & 55.65 & 41.45 & 44.67  \\
         2 & NoStructure*  & 60.45 & 47.99 & 50.15 & 31.31 & 24.31 & 25.65 & 56.04 & 44.17 & 46.33  \\
           3 & STRONG* & 58.84 & 52.47 & 52.62 & 30.38 & 26.93 & 27.04 & 54.67 & 48.40 & 48.70 \\
             \midrule
             \multicolumn{9}{l}{\small{\textit{Max output of 512 token}}} \\
             \midrule

           4   & NoStructure & 57.88 & 53.67 & 51.99 & 30.14 & 27.53 & 26.85 & 53.76 & 49.62 & 48.19 \\
          
           5 & STRONG* & 53.33 & 61.99 & 55.74 & 27.13 & 31.52 & 28.33 & 49.61 & 57.53 & 51.80 \\
    \bottomrule
    \end{tabular}
    \caption{The Complete ROUGE results for various models on the CanLII 1049 oracle dataset.}\label{tab:complete_results_appendix}
\end{sidewaystable}

\newpage
\clearpage
\section{Examples of Different System Outputs}\label{appendix:examples}

\subsection{Different Prompts' Effects}\label{appendix:diff_prompt}
In Table \ref{tab:generation_examples_diff_prompt}, we generate multiple summaries according to different prompts using the best-performing STRONG method and set the maximum length of generation at 512 tokens. We find that the outputs follow the structure prompts to a certain degree. For instance, Variant 1 quickly jumped to the reasoning parts after the first two sentences, while Variant 2 started with multiple clear conclusion sentences on the court's decision and the main issues. 
\subsection{Sample Outputs}
In Table \ref{tab:generation_examples}, we show the examples for different methods under the 256 token max generation limit. We further ask three legal experts to rate the different outputs and analyze on the coverage of argumentative roles. We find that SentBS does a good job of stating an issue, but never reaches the conclusion. The NoStructure -- 256 model fails to give a good statement of the issues, and our STRONG -- 256 produces a more coherent and clear presentation. We additionally include the 512-token version of NoStructure and STRONG outputs in Table \ref{tab:generation_examples_512}. Compared to the shorter NoStructure output that does not clearly state the issue, and it also doesn’t reveal how the issue came out, the legal expert reported that the STRONG - 512 version is very clear and comprehensive. He also raised some concerns about the privacy problem of leaking the decedent's full name.

\begin{table*}
\begin{tabular}{p{1.2in}|p{4.5in}}
\toprule
\Centering{Prompt} & \RaggedRight{Summary}\\
\midrule
\multicolumn{2}{l}{\textit{Human Written Summary}} \\
\midrule
\RaggedRight{ Issue | Conclusion | Conclusion | Reason} &
\RaggedRight{At issue was the interim custody of the child and child support. HELD: Interim custody was given to the petitioner with access to the respondent. The respondent was ordered to pay interim child support in the amount of \$193 per month. On an interim custody application, the Court must take into account the status quo.}\\
\midrule 
\multicolumn{1}{l}{\textit{Generations}} \\
\midrule
\RaggedRight{ Issue | Conclusion | Conclusion | Reason} &
\RaggedRight{At issue was the interim custody of the child and child support.
The parties began cohabitating in May 1998 and separated in April 2000.HELD: The petitioner was awarded interim custody with reasonable access including every second weekend, reasonable telephone access and reasonable access during holiday periods.
Both parents were ordered to complete the Parenting After Separation course.
There was an unhealthy degree of constant bickering.
}\\
\midrule
\RaggedRight{ \textit{Variant 1:} Issue | Conclusion | {Reason | Reason | Reason}} &
\RaggedRight{The petitioner was awarded interim custody with reasonable access including every second weekend, reasonable telephone access and reasonable access during holiday periods.
Both parents were ordered to complete the Parenting After Separation course.
The parties raised in their material the issue of each other's lack of cooperation and communication.
While the Court was not in position to decide whether the accusations of each party were substantiated, it could say with relative certainty that there was an unhealthy degree of constant bickering.
It was evident that as long as the parties work on establishing better communication, the child would benefit from contact with both parents with minimal disruption.
}\\
\midrule 
\RaggedRight{ \textit{Variant 2:}  Conclusion | Issue | Issue | Reason | Reason} &
\RaggedRight{Interim custody of the child was awarded to the mother.
The father was ordered to pay interim child support of \$193 per month.
Both parents were ordered to complete the Parenting After Separation course.HELD: The mother was awarded interim custody with reasonable access including every second weekend, reasonable telephone access and reasonable access during holiday periods.
Interim child support was ordered in the amount of \$125 per month and the father was to pay his proportionate share of childcare costs.
There was an unhealthy degree of constant bickering.
It was evident that as long as the parties worked on establishing better communication, the child would benefit from contact with both parents with minimal disruption.
}\\
\bottomrule
\end{tabular}
\caption{An example of the difference between generated summaries based on different prompts using our best structure prompt model STRONG. The original legal decision id is 2003skqb487.}
\label{tab:generation_examples_diff_prompt}
\end{table*}

\begin{table*}
\small
\begin{tabular}{p{6in}}
\toprule 
 \textbf{ORACLE} \\
\RaggedRight{Statutes – Interpretation – Limitation of Civil Rights Act, Section 2 The plaintiff bank applied by notice of application for a deficiency judgment against the defendant pertaining to the non-purchase money portion of the outstanding indebtedness to the plaintiff pursuant to her mortgage and to some other costs related to the mortgage. The original mortgage executed by the plaintiff was for \$191,000 of which \$176,000 was advanced to purchase her condominium and \$15,000, eight per cent of the total, was advanced for other purchases. The property sold for \$156,000 pursuant to a judicial sale. The sale proceeds were distributed to pay outstanding property taxes, commission, and the principal and interest due under the mortgage in the amount of \$141,945 and the net sale proceeds were to be distributed pro rata. As at November 1, 2017, the principal and interest owing on the non-purchase money portion of the mortgage was \$21,250. The plaintiff requested judgment for the deficiency in the amount of \$9,900 with interest to the date of judgment. It claimed it was entitled to solicitor-client costs and other costs accrued by it under the mortgage. The issue was whether such costs should be subject to the pro rata calculation for the nonpurchase portion of the mortgage, that is, 8 percent of the total costs incurred.  HELD: The application was granted. The court found that the plaintiff was entitled to a deficiency judgment for principal and interest owed under the mortgage in the amount requested. It was also entitled to judgment against the mortgagee for eight percent of the solicitor-client costs, condominium fees and property management charges.}\\
\midrule
\RaggedRight{\textbf{SentBS}}  \\
\RaggedRight{Statutes – Interpretation – Condominium Property Act, 1993, Section 63 The plaintiff bank applied by notice of application for a deficiency judgment against the defendant, Partridge, pertaining to the non-purchase money portion of the outstanding indebtedness to the plaintiff pursuant to her mortgage.
Partridge had originally executed a mortgage for \$191,000, of which \$176,000 was advanced to purchase her condominium unit and \$15,000 for other purchases.
The non-Purchase money amount of the loan advanced and secured by the mortgage was 8
RBC applied for solicitor-client costs and other costs accrued by RBC under the mortgage.
The issue was whether the pro rata distribution applied solely to the indebtedness of principal and accrued interest, or applied to all costs of outstanding mortgage, including solicitor and client costs permitted under the mortgages.
In its application, RBC had calculated the proceeds from the judicial sale that must be distributed pro ratus between the purchase money and non-paid balances owing, resulting in the deficiency of principal.
For the deficiency amount to be set at the rate of 2.}\\
\midrule
\RaggedRight{ \textbf{{NoStructure -- 256}}}\\
\RaggedRight{The petitioner was awarded interim custody with reasonable access including every second weekend, reasonable telephone access and reasonable access during holiday periods.
Both parents were ordered to complete the Parenting After Separation course.
The parties raised in their material the issue of each other's lack of cooperation and communication.
While the Court was not in position to decide whether the accusations of each party were substantiated, it could say with relative certainty that there was an unhealthy degree of constant bickering.
It was evident that as long as the parties work on establishing better communication, the child would benefit from contact with both parents with minimal disruption.
}\\
\midrule 
\RaggedRight{ \textbf{STRONG -- 256}} \\
\RaggedRight{Mortgage – Foreclosure – Determination of InterestThe plaintiff bank applied by notice of application for a deficiency judgment against the defendant, Partridge, pertaining to the non-purchase money portion of the outstanding indebtedness to it pursuant to her mortgage.
The plaintiff also applied to have its lawyers’ costs assessed on solicitor-client basis.
Partridge had originally executed a mortgage for \$191,000, of which \$176,000 was advanced to purchase her condominium unit and \$15,000 for other purchases.
Thus, the mortgage was 8 percent of the total mortgage.HELD: The plaintiff’s application was granted and the plaintiff was granted judgment against Partridge in the amount of \$9,901.23 with interest from November 1, 2017, to the date of judgment at the rate of 2.79 per cent per annum.
It was entitled to solicitor and other costs accrued by the plaintiff under its mortgage.}\\
\bottomrule
\end{tabular}
\caption{Different models' outputs for the legal decision (id: 5\_2018skqb216). The structure prompt is ``Non\_IRC | Non\_IRC | Non\_IRC | Non\_IRC | Non\_IRC | Issue | Issue | Issue | Conclusion | Conclusion | Conclusion ''. }
\label{tab:generation_examples}
\end{table*}

\begin{table*}
\begin{tabular}{p{6in}}
\toprule 
\RaggedRight{ \textbf{{NoStructure -- 512}}}\\
\RaggedRight{Mortgages – Foreclosure – DischargeThe plaintiff applied by notice of application for a deficiency judgment against the defendant pertaining to the non-purchase money portion of the outstanding indebtedness to the plaintiff pursuant to her mortgage.
The plaintiff also applied to have its lawyers’ costs assessed on solicitor-client basis.HELD: The plaintiff was granted judgment in the amount of \$9,901.23 with interest from November 1, 2017 to the date of judgment at the rate of 2.79 per cent per annum for the deficiency of principal and interest owed under the mortgage.
}\\
\midrule 
\RaggedRight{ \textbf{STRONG -- 512}} \\
\RaggedRight{Mortgage – Foreclosure – Pro RataDebtor and Creditor – Mortgage – Non-purchase MoneyThe plaintiff, Royal Bank of Canada, applied by notice of application for a deficiency judgment against the defendant, Doreen Pearl Partridge, pertaining to the non-payment of outstanding indebtedness to the plaintiff pursuant to her mortgage.
The property sold pursuant to an Order Nisi for Sale by Real Estate Listing for \$156,000 was distributed to pay outstanding property taxes, real estate commission and the principal and interest due under the mortgage in the sum of \$141,945.36.
At issue was whether the pro rata distribution applied solely to the indebtedness of principal and accrued interest, or applied to all costs of the outstanding mortgage, including solicitor-client costs permitted under it.
Partridge had originally executed a mortgage for \$191,000 of which \$176,000 advanced to purchase her condominium unit and \$15,000 for other purchases.
Thus, the mortgage was 8 percent of the total mortgage.HELD: The plaintiff was granted judgment against Partridge in the amount of \$9,901.23 with interest from November 1, 2017 to the date of judgment at the rate of 2.79 per cent per annum.
It was entitled to solicitor and client costs and other costs accrued by the plaintiff under its mortgage, that is, 8 per cent of its total outstanding mortgage costs incurred.
Section 63 of The Condominium Property Act, 1993 allows the condominium corporation to register a lien against the title of the unit for unpaid contributions to the common expense fund or the reserve fund.
Secondly, the plaintiff claimed \$1,461.92 for its payment of property management charges for securing and caring for the property, appraisal fee and utilities.
These charges were permitted by s. 8(1) of the Limitation of Civil Rights Act (LCRA) and any inspections and administration fees had not been claimed by RBC.
Further, the property management charge was recoverable under the terms of the mortgage.}\\
\bottomrule
\end{tabular}
\caption{NoStructure and Strong models' outputs for the legal decision (id: 5\_2018skqb216) under 512 max length generations. The structure prompt is ``Non\_IRC | Non\_IRC | Non\_IRC | Non\_IRC | Non\_IRC | Issue | Issue | Issue | Conclusion | Conclusion | Conclusion ''. }
\label{tab:generation_examples_512}
\end{table*}

\section{Human Evaluation Details}\label{appendix:human_evaluate}
We conducted evaluations with a total of three legal experts, all of whom hold a J.D. degree and possess a minimum of four years of experience in providing professional legal services. The experts were assigned five randomly sampled legal cases, each accompanied by the oracle reference summary, as well as the generated outputs from the following five models: (1) SentBS with a length of 256 tokens, (2) NoStructure with a length of 256 tokens, (3) STRONG with a length of 256 tokens, (4) NoStructure with a length of 512 tokens, and (5) STRONG with a length of 512 tokens. The experts were presented with the reference summary and all five system outputs in the same row of an Excel file. They were then asked to provide reflections on the faithfulness and coherence of each system output while considering the inclusion of essential argument roles components such as Issue, Reason, and Conclusion compared to the reference summary. Given that the instruction does not specifically inquire about the ranking nor ask evaluators to provide numerical scores,  the primary author instead offers an interpretation of the free-text reflections by conducting comparative analyses across various outputs and allocating a relative ranking. We release all reflections for further studies in \url{https://github.com/cs329yangzhong/STRONG}. Table \ref{tab:human_evaluation_example} shows one example of evaluators' reflections on a case, and Table \ref{tab:author_evaluate_number} shows the author's ranking interpretation.

\begin{table*}
\centering
\begin{tabular}{p{1.2in}|p{3in}}
\toprule
\RaggedRight{Model Output} & \RaggedRight{Annotator Reflection} \\
\midrule
\RaggedRight{SentBS} &
\RaggedRight{Does a good job stating an issue, but never reaches the conclusion.}\\
\midrule 
\RaggedRight{NoStructure -- 256} &
\RaggedRight{No good statement of the issue, but maybe readers could infer the issue based on the conclusion “It was entitled to solicitor and other costs accrued by the plaintiff under its mortgage.”
The interest payment isn’t important enough to be in the summary} \\
\midrule
\RaggedRight{STRONG -- 256} &
\RaggedRight{Fairly coherent, but it’s not totally clear the dispute is about “solicitor and client costs permitted under the mortgages.} \\ 
\midrule
\RaggedRight{NoStructure -- 512} &
\RaggedRight{It’s not very clear about the issue, and it also doesn’t reveal how the issue came out.} \\
\midrule
\RaggedRight{STRONG -- 512} &
\RaggedRight{This is very clear. Using the defendant’s full name might be a privacy problem. I also wonder if there’s a copyright problem with using the subject classification system at the start of the summary. It looks like it’s from the Law Society of Saskatchewan.} \\
\bottomrule
\end{tabular}
\caption{A sample of the human evaluation on different model outputs. It corresponds to Annotator 3's reflection for the second legal decision summary group in Table \ref{tab:author_evaluate_number}.}
\label{tab:human_evaluation_example}
\end{table*}

\begin{table*}[t!]
\small
    \centering
    \renewcommand{\arraystretch}{1}
    \begin{tabular}{l|c|c|c|c|c}
    
   \toprule
   \textbf{Annotator} & \textbf{SentBS} & \textbf{NoStructure-256} & \textbf{STRONG-256} &\textbf{NoStructure-512} & \textbf{STRONG-512} \\
   \midrule
    Anno. 1 & 5 & 1 & 3 & 1 & 4 (too detailed)\\
     Anno. 2 &\multicolumn{5}{c}{N/A} \\
      Anno. 3 & 3 (fluency problem) & 1 & 3 (same as sentbs) & 2 & 5\\
      \midrule
    Anno. 1 &  5 (no conclusion) & 4 (no issue) & 3 & 2  & 1 (fairly clear)\\
     Anno. 2 & 5 (lack conclusion) & 2 & 4 (factual errors) &2 (lack conclusion) & 1 (very good)\\
      Anno. 3 & 5 (never concludes) & 3 (good) & 3 & 2 (was nice, but ...) & 1 (great)\\
      \midrule
    Anno. 1 & \multicolumn{5}{c}{N/A} \\
     Anno. 2 &  5 (wrong issue) & 1  & 3 (too many details) &  4 (erratic contents) & 2 (too detailed)\\
      Anno. 3 & 5 (very confusing) & 1 &  3 (bad grammar) & 4 (not reliable)    & 2 \\
      \midrule
    Anno. 1 & 3 & 3 & 5 &  1 & 2\\
     Anno. 2 & 4 (no I, C) & 4 (same to SentBS) & 3 (no issue) & 1 & 2 (fairly good)\\
      Anno. 3 & \multicolumn{5}{c}{N/A} \\
      \midrule
    Anno. 1 & \multicolumn{5}{c}{N/A}  \\
     Anno. 2   & 2 (not good) & 2 (same to SentBS) & 2 (same to SentBS) & 1 & 5 (bad summary)\\
      Anno. 3 & 3 (generally good) & 3  & 2 & 1 & 5 (bad summary)\\
      
    \bottomrule
    \end{tabular}
    \caption{The inferred rankings of different system outputs, determined based on human reflections over five legal decision summaries. Some annotators did not annotate a specific summary, and the row is represented by ``N/A''.}
    \label{tab:author_evaluate_number}
\end{table*}

\label{sec:appendix}

\end{document}